\documentclass[lettersize,journal]{IEEEtran}
\usepackage{amsmath,amsfonts}
\usepackage{bm}
\usepackage{algorithmic}
\usepackage{algorithm}
\usepackage{array}
\usepackage[caption=false,font=normalsize,labelfont=sf,textfont=sf]{subfig}
\usepackage{textcomp}
\usepackage{stfloats}
\usepackage{url}
\usepackage{verbatim}
\usepackage{graphicx}
\usepackage{cite}
\usepackage{multirow}
\newcommand{\thickhline}{\noalign{\hrule height 1.0pt}}

\begin{document}

\title{NeRT: Implicit Neural Representations for General Unsupervised Turbulence Mitigation}
\author{Weiyun Jiang, Yuhao Liu, Vivek Boominathan, and Ashok Veeraraghavan,~\IEEEmembership{Fellow,~IEEE}
        % <-this % stops a space
\thanks{Note: A much smaller version of this paper was submitted for CVPR Workshop. We have uploaded that version as supplementary for full disclosure.}% <-this % stops a space
%\thanks{Manuscript received April 19, 2021; revised August 16, 2021.}
}

\markboth{Journal of \LaTeX\ Class Files,~Vol.~14, No.~8, August~2021}%
{Shell \MakeLowercase{\textit{et al.}}: A Sample Article Using IEEEtran.cls for IEEE Journals}

\IEEEpubid{0000--0000/00\$00.00~\copyright~2021 IEEE}

\maketitle
\begin{abstract}
The atmospheric and water turbulence mitigation problems have emerged as challenging inverse problems in computer vision and optics communities over the years. However, current methods either rely heavily on the quality of the training dataset or fail to generalize over various scenarios, such as static scenes, dynamic scenes, and text reconstructions. We propose a general implicit neural representation for unsupervised atmospheric and water turbulence mitigation (NeRT). NeRT leverages the implicit neural representations and the physically correct tilt-then-blur turbulence model to reconstruct the clean, undistorted image, given only dozens of distorted input images. Moreover, we show that NeRT outperforms the state-of-the-art through various qualitative and quantitative evaluations of atmospheric and water turbulence datasets. Furthermore, we demonstrate the ability of NeRT to eliminate uncontrolled turbulence from real-world environments. Lastly, we incorporate NeRT into continuously captured video sequences and demonstrate $48 \times$ speedup. 
\end{abstract}

\section{Introduction}
\label{sec:intro}
\IEEEPARstart{G}{eneral} turbulence distortion is a common phenomenon that we encounter in our daily lives. Whenever light waves pass through nonuniform mediums before reaching the imaging system, they are refracted and distorted. In a uniform medium, such as free space and outer space, the imaging system always captures clean and sharp images. However, when light passes from a dense medium to a sparse medium, it is refracted according to Snell's law, resulting in a sequence of distorted images. Seeing through water is a simple example of such distortion, where water and air act as two mediums with different refractive indices. When the water surface moves up and down randomly, it becomes challenging to mitigate water turbulence and reconstruct the original undistorted scene underwater. In addition, seeing through atmospheric turbulence presents an even more challenging task, especially for ground-based long-range imaging systems. The intermediate medium between the objects and the imaging systems may extend for several kilometers, and various factors, such as wind, humidity, temperature, and distance, can affect the medium's refractive index along the light path. Unlike water turbulence, where we only need to consider the water-air interface, atmospheric turbulence involves an unlimited number of nonuniform interfaces that refract light arbitrarily, resulting in spatiotemporally varying tilting and blurring in distorted images~\cite{roggemann1996imaging}.

Existing deep learning approaches~\cite{feng2023turbugan, zhang2022imaging, mao2020image, jin2021neutralizing, gao2019atmospheric, nieuwenhuizen2019deep, li2018learning} for seeing through atmospheric turbulence require a huge amount of training dataset. In the first place, it is a challenging task to build a physically accurate and fast simulator~\cite{chimitt2020simulating, mao2021accelerating, chimitt2022real, hardie2017simulation}. Additionally, by leveraging domain-specific priors, these existing deep learning approaches inevitably have dataset biases~\cite{torralba2011unbiased, cole2022learned, zhang2022fedaudio} and poor performance for out-of-domain distributions. 

Classical non-deep learning approaches~\cite{mao2020image, shimizu2008super, anantrasirichai2018atmospheric, caliskan2014atmospheric} for atmospheric turbulence mitigation do not require a huge amount of training dataset. However, they need to rely on some non-rigid registration techniques, such as B-spline functions~\cite{shimizu2008super}, optical flow~\cite{mao2020image, caliskan2014atmospheric}, and diffeomorphism~\cite{gilles2008atmospheric}, to model the grid deformation under atmospheric turbulence. All of these non-rigid registration methods, such as B-spline functions, optical flow, and diffeomorphism, require a reference frame to start with. The reference frames selected by these methods do not accurately represent the original sharp image. Mao et al.~\cite{mao2020image} select the sharpest distorted input frame, which still contains blurring and tilting, as the reference frame for the optical flow algorithm. Additionally, Shimizu et al.~\cite{shimizu2008super} select the temporal average of all the distorted input frames, which also contain noise, blurring, and tilting, as the reference frame. Furthermore, Gilles et al.~\cite{gilles2008atmospheric} choose the reference frame as the temporal median of all the distorted input frames, which is inherently distorted by noise, blurring, and tilting. 

\begin{figure*}[!ht]
	\centering
		\includegraphics[width=7.5in]{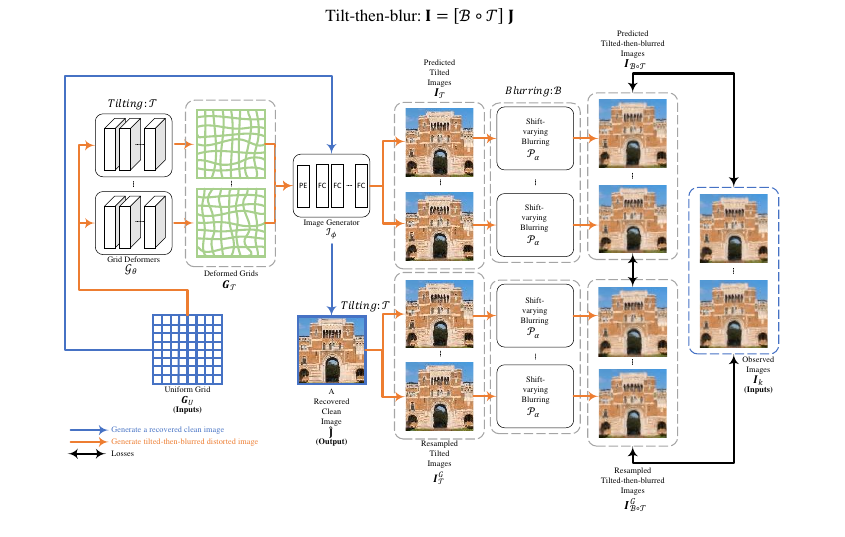} 
\caption{The overall architecture of NeRT. The network predicts a clean and sharp image $\hat{\mathbf{J}}$, given a series of observed atmospheric turbulence distorted images $\mathbf{I}$. We compute $L_1$ loss between predicted tilted-then-blurred images $\mathbf{I}_{\mathcal{B}\circ \mathcal{T}}$, resampled tilted-then-blurred images $\mathbf{I}^{G}_{\mathcal{B}\circ \mathcal{T}}$, and observed images $\mathbf{I}$ during optimization to update the parameters in image generators $\mathcal{I}_{\phi}$, grid deformers $\mathcal{G}_{\theta}$, and shift-varying blurring $\mathcal{P}_{\alpha}$.}
	\label{fig:architecture}
\end{figure*}

To address the above challenges, we primarily focus on atmospheric turbulence and design an implicit neural representation for unsupervised atmospheric mitigation (NeRT), removing temporally and spatially tilting and blurring. Building upon this success, we also extend NeRT to address other types of turbulence distortions, including mitigating water turbulence and handling water ripple reflection turbulence in uncontrolled environments. The key idea is to constrain the network to learn the physically correct air turbulence forward model, the tilt-then-blur model~\cite{chan2022tilt}. Inspired by NDIR~\cite{li2021unsupervised}, NeRT can model temporally and spatially tilting using deformed grids and implicit neural representations. For instance, if the implicit neural representations take uniform undistorted coordinates, they render clean and sharp images. If the implicit neural representations take distorted coordinates, they will output the corresponding distorted images under atmospheric turbulence. 
\IEEEpubidadjcol
The overall architecture of our learning framework, NeRT, is depicted in Figure~\ref{fig:architecture}. The network contains three major components, grid deformers $\mathcal{G}$ that estimate the spatially and temporally varying tilting at each pixel location, an image generator $\mathcal{I}$ that outputs pixel values at corresponding coordinates, and shift-varying blurring $\mathcal{P}$ that approximates the spatially blurring at each pixel location. 

We perform extensive experiments on both atmospheric and water turbulence datasets. Our study demonstrates that NeRT outperforms the state-of-the-art supervised methods in mitigating atmospheric turbulence. In addition, NeRT performs comparably to the state-of-the-art methods in removing water turbulence. We further demonstrate NeRT's robustness in uncontrolled environments, where it effectively mitigates water ripple reflection turbulence. Our specific contributions include as follows: 
\begin{itemize}%[leftmargin=*]
\item We are the first to propose an unsupervised and physically grounded deep learning method for general turbulence mitigation. The pipeline follows the physically correct forward turbulence model --- the tilt-then-blur model\cite{chan2022tilt}. 
\item Our unsupervised algorithm is highly generalizable as it can recover clean and distortion-free images without domain-specific priors such as distorted-clean image pairs. We validate this generalizability by testing on image datasets seeing through atmospheric and water turbulence. 
\item We test our method on newly collected videos of water reflection turbulence in uncontrolled environments and further validate our method's generalizability.
% \item We successfully deploy our method on real-time continuously captured video footage and achieve rapid convergence within $10$ seconds on the latest frame.
\item Finally, we show a path toward faster and live reconstruction by processing the current video frame based on network parameters learned from previous frames. 
\end{itemize}

\section{Related work}
\label{sec:related}
\noindent \textbf{Implicit neural representations.} Implicit neural representations, which use multi-layer perceptions (MLPs) as the backbone networks, store 2D images~\cite{sitzmann2020implicit, tancik2020fourier} and 3D shapes~\cite{mildenhall2021nerf, martin2021nerf} as continuous functions. The inputs of implicit neural representations are $2$D or $3$D coordinates, while the outputs are the corresponding signal. This kind of continuous representation shows not only extraordinary results in overfitting a single image or multiple images but also exceeds other state-of-the-art architectures in solving inverse problems, such as single-image superresolution~\cite{chen2021learning}, medical image reconstruction~\cite{shen2022nerp} and medical image registration~\cite{wolterink2022implicit}. Our work, NeRT, uses implicit neural representations as $2$D image functions to render distorted images under atmospheric turbulence and clean, undistorted images.

\noindent \textbf{Atmospheric turbulence mitigation.} Many works have been proposed to undistort the effects of atmospheric turbulence. Some recent works have demonstrated the uses of transformer architectures as supervised turbulence removal networks for single frame~\cite{mao2022single, zamir2021multi, zamir2022restormer, park2020multi, yasarla2021learning, mei2023ltt, nair2023ddpm} and muti-frame~\cite{zhang2022imaging, chan2022basicvsr++, jin2021neutralizing, liang2022vrt, oreifej2012simultaneous} atmospheric turbulence mitigation tasks. These proposed supervised transformer architectures must rely on fast and physically accurate simulators to generate a huge collection of paired distorted-clean image pairs for training datasets. Although they have fast inference speed, they are hard to generalize over out-of-the-domain datasets. Our proposed unsupervised learning architecture, NeRT, does not require any datasets for pretraining and, thus, can generalize over all kinds of datasets. TurbuGAN~\cite{feng2023turbugan} proposes a self-supervised approach for imaging through turbulence by leveraging an adversarial learning framework and a fast turbulence simulator~\cite{mao2021accelerating}. This approach requires no paired training datasets; however, it is hard to generalize domain-specific priors over out-of-the-domain distributions. NDIR~\cite{li2021unsupervised} is the closest to our work. This method exploits convolutional neural networks to model non-rigid distortion. However, it relies on an off-the-shelf physically incorrect spatially invariant deblurring algorithm~\cite{xu2013unnatural}. Our method, NeRT, incorporates a physically grounded, spatially and temporally varying deblurring approach to restoring the sharp image.  

\noindent \textbf{Seeing through water.} Over the years, numerous methods have been proposed to restore the clear underwater scene using multiple images or a single image, distorted by water turbulence. Tian et al.~\cite{tian2009seeing} propose to build a compact spatial distortion model of the water surface using the wave equation. Together with a novel tracking technique, they propose to use the compact model to remove water turbulence. Oreifej et al.~\cite{oreifej2011two} presents a two-stage approach for seeing through the water. During the first stage, they simultaneously estimate the spatially-varying blur kernel and non-rigid distortion to de-warp the distorted frames. During the second stage, they apply an off-the-shelf low-rank optimization algorithm to denoise the registered frames. Li et al.~\cite{li2018learning} propose to train a deep convolution neural network to undistort the dynamic refractive effects of water turbulence using only a single image. Given the supervised method requires only a single image and thousands of carefully collected training image pairs, Li et al.~\cite{li2018learning} have a very poor ability for generalization.

\section{Physically grounded restoration network}
\label{sec:network}
Given dozens of turbulence-distorted input images, $\mathbf{I} \in \mathbb{R}^{m\times n \times 3}$, NeRT predicts a clean and sharp image without any distortions, $\hat{\mathbf{J}} \in \mathbb{R}^{m \times n \times 3}$ without any domain-specific image priors. 

The key idea is to optimize an unsupervised model to precisely represent both the spatially and temporally varying tiltings and the spatially and temporally varying blurrings from general turbulence, such as water and atmospheric turbulence. Once we have the ability to estimate the forward model, reversing it becomes a simple task that allows us to reconstruct the undistorted image, denoted as $\hat{\mathbf{J}}\in \mathbb{R}^{m \times n \times 3}$. 

Our model has three major components, \textbf{grid deformers}, \textbf{image generators}, and \textbf{shift-varying blurring}. Each grid deformer, $\mathcal{G}_{\theta}$ aims to estimate the spatially varying and temporally varying tiltings (non-rigid deformations) of every distorted input image. Then, given the estimated non-rigid deformed $2$D pixel coordinates, the image generator, $\mathcal{I}_{\phi}$ generates the tilted image $\mathbf{I}_{\mathcal{T}} \in \mathbb{R}^{m \times n \times 3}$. At the same time, given the uniform and undistorted $2$D pixel coordinates, $\mathbf{G}_U \in \mathbb{R}^{m \times n \times 2}$, the image generator, $\mathcal{I}_{\phi}$ generates the clean and sharp image $\hat{\mathbf{J}} \in \mathbb{R}^{m \times n \times 3}$. We also resample the clean and sharp image $\hat{\mathbf{J}} \in \mathbb{R}^{m \times n \times 3}$ based on the estimated deformed $2$D pixel coordinates, $\mathbf{G}_\mathcal{T} \in \mathbb{R}^{m\times n \times 2}$ to predict the resampled tilted image $\mathbf{I}^{G}_{\mathcal{T}} \in \mathbb{R}^{m \times n \times 3}$. Finally, given generated tilted image $\mathbf{I}_{\mathcal{T}} \in \mathbb{R}^{m \times n \times 3}$ and resampled tilted image $\mathbf{I}_{\mathcal{T}} \in \mathbb{R}^{m \times n \times 3}$, the shift-varying blurring predicts the generated tilted-then-blurred image $\mathbf{I}_{\mathcal{B}\circ \mathcal{T}} \in \mathbb{R}^{m \times n \times 3}$ and resampled tilted-then-blurred image $\mathbf{I}^G_{\mathcal{B} \circ \mathcal{T}} \in \mathbb{R}^{m \times n \times 3}$ respectively.

One crucial aspect of NeRT is its unsupervised optimization of grid reformers, $\mathcal{G_{\theta}}$, the image generator, $\mathcal{I}_{\phi}$, and shift-varying blurring, $\mathcal{P}_{\alpha}$, which enables the identification of perfect spatiotemporal blurring kernels and deformed grids for each distorted input frame. Section~\ref{sec:exp} showcases NeRT's superiority over other state-of-the-art methods in effectively mitigating various types of turbulence, including atmospheric and water turbulence. Furthermore, Section~\ref{sec:wild} presents qualitative results that demonstrate the robustness of NeRT in eliminating uncontrolled turbulence from real-world environments.
\subsection{Forward atmospheric turbulence model}

Imagine the light reflected from a scene, represented as a clean and sharp image $\mathbf{J}$, travels through space with a spatially and temporally varying index of refraction. The light finally arrives at a passive imaging device, such as a digital single-lens reflex (DSLR) camera, forming many distorted images $\mathbf{I}$ over time. Each $\mathbf{I}$ has stochastic distortion at different pixel locations and time stamps. The generalized forward atmospheric turbulence model can be written as~\cite{chan2022tilt}:
\begin{equation}
    \mathbf{I}(x,y,t) =\mathcal{H}_t(\mathbf{J}(x,y,t)),
\end{equation}
where $\mathcal{H}$ is a general linear distortion operator. However, we desire to decompose $\mathcal{H}$ into simpler operations for the computational tractability of inverting the distortion. Fortunately, representation using Zernike bases allows us to do so.

The distortion from atmospheric turbulence can be parameterized by coefficients of Zernike polynomials in the phase space~\cite{chimitt2020simulating}. The representation in the Zernike space allows us to decouple the distortion into interpretable operations of tilting operation ($\mathcal{T}$) and blurring operation ($\mathcal{B}$). The tilt $\mathcal{T}$ is encoded by the first two Zernike bases (barring the constant term) and is the most significant contributor to shifting the center of mass of the distortion spread. The blur $\mathcal{B}$ is encoded by the remaining Zernike bases and describes the distortion spread. The question is how to compose the decoupled operations to describe the actual atmospheric distortion accurately. There are two possible options:

\begin{itemize}
\centering
\setlength\itemsep{.01em}
    \item[] Blur-then-tilt: $\mathbf{I} = [\mathcal{T} \circ \mathcal{B}]\mathbf{J}$, 
    \item[] Tilt-then-blur: $\mathbf{I} = [\mathcal{B} \circ \mathcal{T}]\mathbf{J}$,
\end{itemize}  
where $\circ$ is a functional composition operator, and the composition is read from right to left.
\begin{figure}[!t]
	\centering
		\includegraphics[width=3.5in]{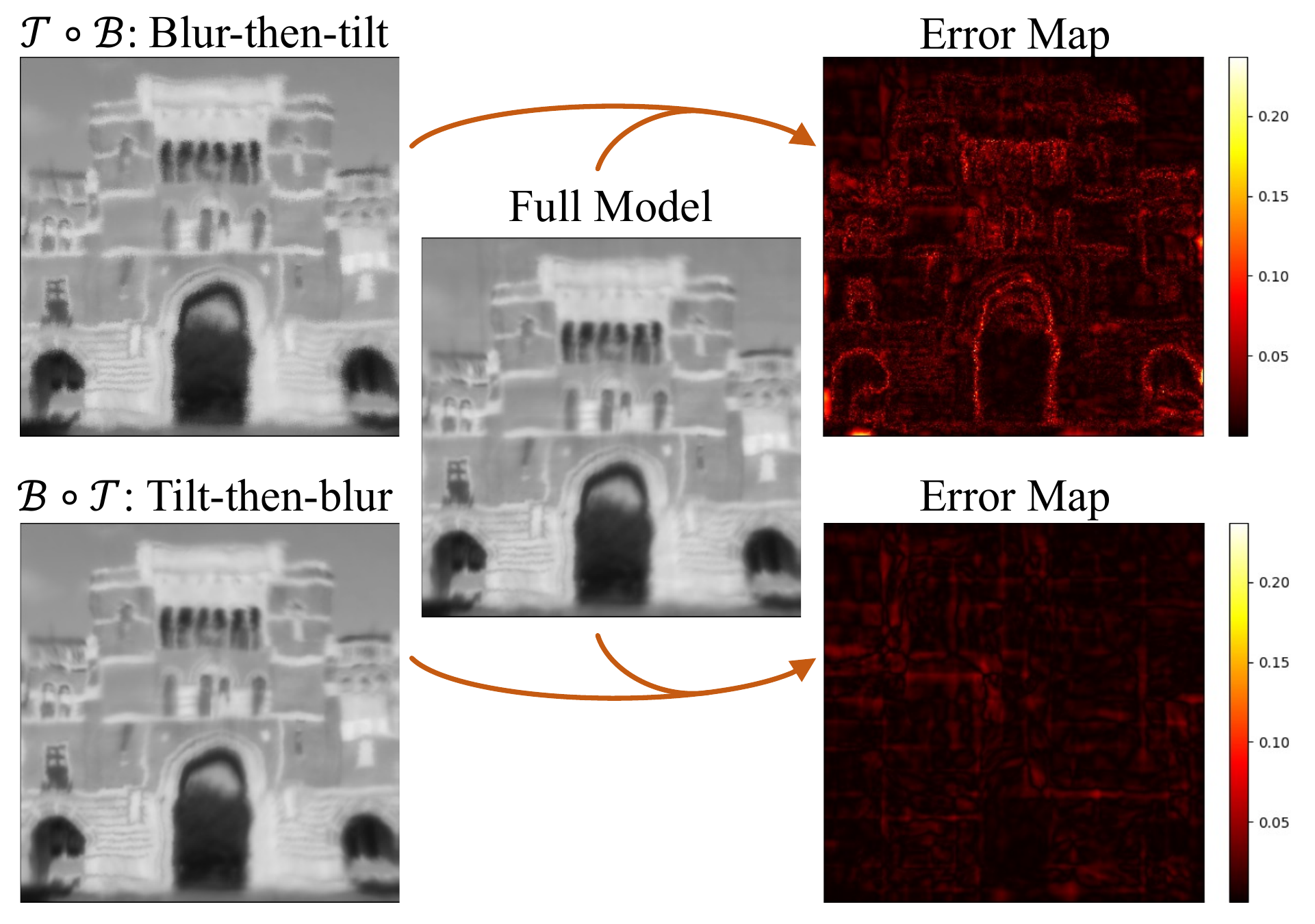} 
\caption{Simulation of turbulence using $\mathcal{B} \circ \mathcal{T}$, $\mathcal{T} \circ \mathcal{B}$, and the full model, along with their respective error maps with respect to the full model. The error of the blur-then-tilt model, $\mathcal{T} \circ \mathcal{B}$ occurs at edges and high-resolution regions.} 
	\label{fig:why}
\end{figure}

\begin{figure*}[!hb]
	\centering
		\includegraphics[width=7.2in]{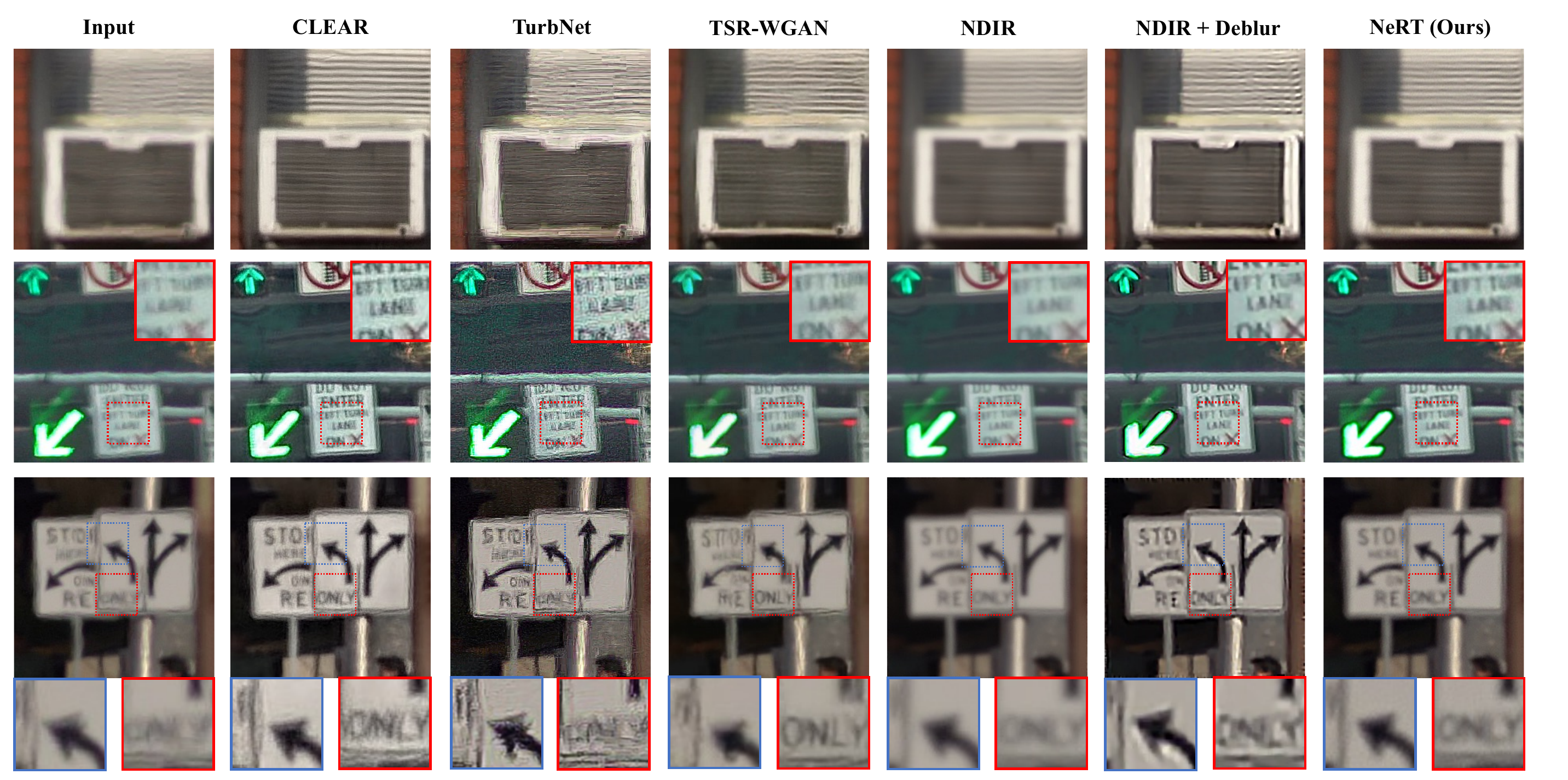} 
\caption{Qualitative results from our newly captured in-the-wild static scene atmospheric turbulence datasets. All sequences are captured using a Sony a7 III mirrorless camera with a 400 mm lens. TurbNet~\cite{mao2022single} and TSR-WGAN~\cite{jin2021neutralizing} are supervised SOTA while CLEAR~\cite{anantrasirichai2018atmospheric} and NDIR~\cite{li2021unsupervised} are unsupervised SOTA. NDIR+Deblur post-processes the NDIR results by applying an off-the-shelf deblurring algorithm~\cite{li2018learning}, which has no domain-specific knowledge of atmospheric turbulence.}
\label{fig:static_own}
\end{figure*}
Many of the previous works~\cite{anantrasirichai2018atmospheric, mao2020image, zhu2012removing, gilles2008atmospheric} opt to use the blur-then-tilt model, whose inversion is to untilt first and then to deblur. They choose this route because it is relatively easier to estimate the tilt first using well-known non-rigid registration computations such as optical flow~\cite{mao2020image, caliskan2014atmospheric}, diffeomorphism~\cite{gilles2008atmospheric}, and B-spline function~\cite{shimizu2008super}. Then, after untilting, an off-the-shelf deblurring algorithm is used to deblur~\cite{li2021unsupervised}.

A recent study~\cite{chan2022tilt} shows, by careful analysis, that the blur-then-tilt model is inaccurate and that the tilt-then-blur model is physically more accurate. However, they still remarked that when it comes to natural images, the approximate blur-then-tilt model is sufficient because of its ease of parameter estimation. In our work, we make the case with a similar analysis that the errors of the approximate blur-then-tilt model accumulate at high-frequency edges and thereby using the correct tilt-then-blur model is important when reconstructing images at high fidelity. We use the correct model of tilt-then-blur to represent our forward model, which is given as follows~\cite{chan2022tilt}:
\begin{equation}
\label{eqn:forward}
    \mathbf{I}(x,y,t) = [\mathcal{B} \circ \mathcal{T}]\mathbf{J}(x,y,t),
\end{equation}
where $\mathcal{B}$ denotes a temporally and spatially varying blurring operator, $\mathcal{T}$ represents a temporally and spatially varying tilting operator and $\circ$ is a functional composition operator. We apply spatially and temporally varying tilting operators to the clean image $\mathbf{J}(x,y)$ first to obtain multiple tilted images $\mathbf{I}_\mathcal{T}$ at different time stamps. Then, we apply spatially varying blurring operators to those tilted images $\mathbf{I}_\mathcal{T}$ to render corresponding final tilted-then-blurred images $\mathbf{I}=\mathbf{I}_{\mathcal{B} \circ \mathcal{T}}$ under atmospheric turbulence.

\begin{figure*}[!t]
	\centering
		\includegraphics[width=7.2in]{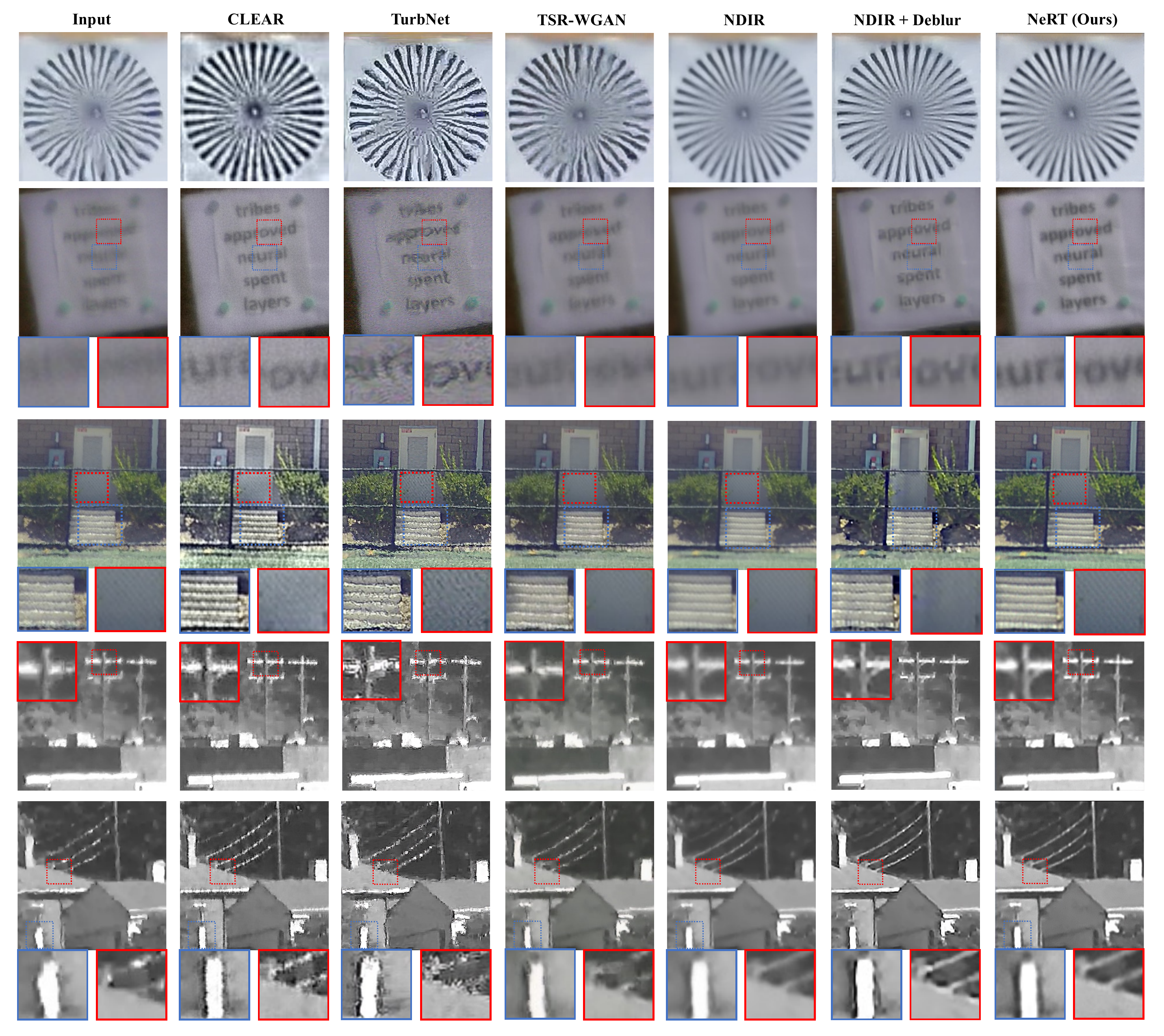} 
\caption{Qualitative results from the static scene atmospheric turbulence datasets, including Siemens star dataset~\cite{gilles2017open}, text dataset~\cite{mao2022single}, door dataset~\cite{gilles2017open} and in-the-wild dataset~\cite{mao2020image}. We compare NeRT with other supervised~\cite{mao2022single} and unsupervised~\cite{li2021unsupervised, anantrasirichai2018atmospheric} SOTA. NeRT is able to achieve high spatial resolution, recover high-contrast text, and reconstruct fine details, such as wire fences. CLEAR~\cite{anantrasirichai2018atmospheric}, TSR-WGAN~\cite{jin2021neutralizing}, and TurbNet~\cite{mao2022single} fails to mitigate the atmospheric turbulence, while NDIR~\cite{li2021unsupervised} fails to preserve fine details of the wire fences and electric poles. NDIR+Deblur relies on an off-the-shelf general-purpose deblurring software~\cite{li2018learning}, which estimates the atmospheric turbulence PSF inaccurately, and produces ringing and noisy artifacts along edges.}
\label{fig:static}
\end{figure*}

\subsection{Why choose tilt-then-blur?}

Although the two compositions $\mathcal{B} \circ \mathcal{T}$ (tilt-then-blur) and $\mathcal{T} \circ \mathcal{B}$ (blur-then-tilt) are analytically different, their impacts on images of natural scenes tend to be similar, and the differences might be imperceptible~\cite{chan2022tilt}. However, the errors in the incorrect $\mathcal{T} \circ \mathcal{B}$ model can quickly accumulate at the edges and high-resolution regions, as shown in the analysis below. 

For simplicity, let us assume that the blurring is spatially invariant. We can write the equations of the two models as~\cite{chan2022tilt}:
\begin{align}
    \mathbf{I}_{\mathcal{T} \circ \mathcal{B} } &= \sum_{j=1}^{N} g(\bm{x}_i-\bm{u}_j)\bm{J}(\bm{u}_j-\bm{t}_i), \\
    \mathbf{I}_{\mathcal{B} \circ \mathcal{T} } &= \sum_{j=1}^{N} g(\bm{x}_i-\bm{u}_j)\bm{J}(\bm{u}_j-\bm{t}_j),
\end{align}
where $g$ is the spatially invariant blur, and $t$ is the tilt. Note that the subtlety is captured in the indexing of $t$. More details regarding the derivation can be found in~\cite{chan2022tilt}.

We may now evaluate the difference between the correct tilt-then-blur model and the incorrect blur-then-tilt model as

\begin{equation} 
\begin{split}
 \mathbf{I}_{\mathcal{T} \circ \mathcal{B} } &- \mathbf{I}_{\mathcal{B} \circ \mathcal{T} } \\
 & = \sum_{j=1}^N g(\bm{x}_i-\bm{u}_j)[\bm{J}(\bm{u}_j-\bm{t}_i) - \bm{J}(\bm{u}_j-\bm{t}_j)]\\
 & \approx \sum_{j=1}^{N} g(\bm{x}_i-\bm{u}_j) \nabla\bm{J}(\bm{u}_j^T)(\bm{t}_i-\bm{t}_j),
\end{split}
\end{equation}

\noindent where $\nabla\bm{J}(\bm{u}_j^T)$ stands for the image gradient, and $\bm{t}_i-\bm{t}_j$ represents the random tilt. For natural scene images, the image gradients $\nabla\bm{J}(\bm{u}_j^T)$ are typically sparse, and the error between the two models is close to zero for most of the image regions. However, the image gradients are strong at edges and high-resolution regions, and there will be a significant error between the two models (Figure~\ref{fig:why}). Thus, it is sub-optimal to solve atmospheric mitigation problems following the incorrect blur-then-tilt model. Using the correct tilt-then-blur model gives us the opportunity to recover edges and high-resolution details from a time window of dynamically distorted frames.

\subsection{Network Structure}

Figure~\ref{fig:architecture} demonstrates the architecture of NeRT. Our model has three major components, grid deformers, image generators, and shift-varying blurring.

\noindent \textbf{Grid deformers} $\mathcal{G_{\theta}}$ take uniform grid $\mathbf{G}_U \in \mathbb{R}^{2 \times m\times n}$ as inputs, where $m$ and $n$ are the image sizes, and output deformed grid, $\mathbf{G}_\mathcal{T} \in \mathbb{R}^{2 \times m\times n}$. Like NDIR~\cite{li2021unsupervised}, each grid deformer $\mathcal{G}_{\theta}$ consists of four convolutional layers with 256 channels and ReLU activation layers. Similarly, we have a dedicated grid deformer $\mathcal{G}_{\theta}$ for each distorted image $\mathbf{I} \in \mathbb{R}^{3 \times m \times n}$.

\noindent \textbf{Image generator} $\mathcal{I_\phi}$ takes deformed $2$D pixel coordinates $\mathbf{G}_{\mathcal{T}} \in \mathbb{R}^{m \times n \times 2}$ as inputs, and output 3D RGB pixel value that corresponds to the tilted images $\mathbf{I}_\mathcal{T} \in \mathbb{R}^{m \times n \times 3}$. It can also take uniform $2$D pixel coordinates $\mathbf{G}_U \in \mathbb{R}^{m \times n \times 2}$ as inputs and output colored pixel value of the clean image $\mathbf{J} \in \mathbb{R}^{m \times n \times 3}$. We build our image generator as an implicit representation consisting of five fully connected layers of hidden size 256 with ReLU activation and positional encoding. We implement our coordinate-based MLPs following previous work, SIREN~\cite{sitzmann2020implicit} and Fourier feature network~\cite{tancik2020fourier}. Specifically, we reshape the 2D pixel coordinates $\mathbf{G} \in \mathbb{R}^{{m \times n \times 2}}$ as $\mathbf{G} \in \mathbb{R}^{m \cdot n \times 2}$ to parse into the coordinate-based MLPs. Additionally, we reshape the output of the coordinate-based MLPs, corresponding 3D RGB pixel value, $\mathbf{J} \in \mathbb{R}^{m \cdot n \times 3}$ as $\mathbf{J} \in \mathbb{R}^{m \times n \times 3}$.

\begin{table*}[!ht]
\centering
\caption{Quantitative performance on synthetic atmospheric turbulence dataset created using simulator~\cite{mao2021accelerating}. $\uparrow$ means the higher the better. We highlight the best results using bold text.}
\begin{tabular}{|c|c|c|c|c|c|c|}
\hline
Strengths    & Metrics   & CLEAR~\cite{anantrasirichai2018atmospheric}   & TurbNet~\cite{mao2022single} &     TSR-WGAN~\cite{jin2021neutralizing} & NDIR~\cite{li2021unsupervised}  & NeRT (Ours)            \\ \thickhline
\multirow{2}{*}{\begin{tabular}[c]{@{}c@{}}Weak\\ ($D/r_0$ = $1.5$)\end{tabular}}   & PSNR $\uparrow$ (dB) & $24.085$          & $21.281$  &  $20.804$ & $25.459$ & $\mathbf{26.133}$ \\ \cline{2-7} 
& SSIM $\uparrow$     & $0.749$  & $0.643$   &  $0.646$ & $0.853$  & $\mathbf{0.871}$           \\ \hline
\multirow{2}{*}{\begin{tabular}[c]{@{}c@{}}Medium\\ ($D/r_0$ = $3$)\end{tabular}}   & PSNR $\uparrow$ (dB) & $21.431$          & $20.160$  &  $18.970$ & $24.302$ & $\mathbf{25.304}$ \\ \cline{2-7} 
& SSIM $\uparrow$      & $0.607$  & $0.567$   &  $0.531$ & $0.816$  & $\mathbf{0.846}$           \\ \hline
\multirow{2}{*}{\begin{tabular}[c]{@{}c@{}}Strong\\ ($D/r_0$ = $4.5$)\end{tabular}} & PSNR $\uparrow$ (dB) & $20.159$          & $19.298$  &  $16.105$ & $23.166$ & $\mathbf{24.006}$ \\ \cline{2-7} 
& SSIM $\uparrow$      & $0.526$ & $0.512$   &  $0.453$ & $0.772$ & $\mathbf{0.801}$           \\ \hline
\end{tabular}
\label{tab:syn}
\end{table*}

\begin{figure*}[!hb]
	\centering
		\includegraphics[width=7.0in]{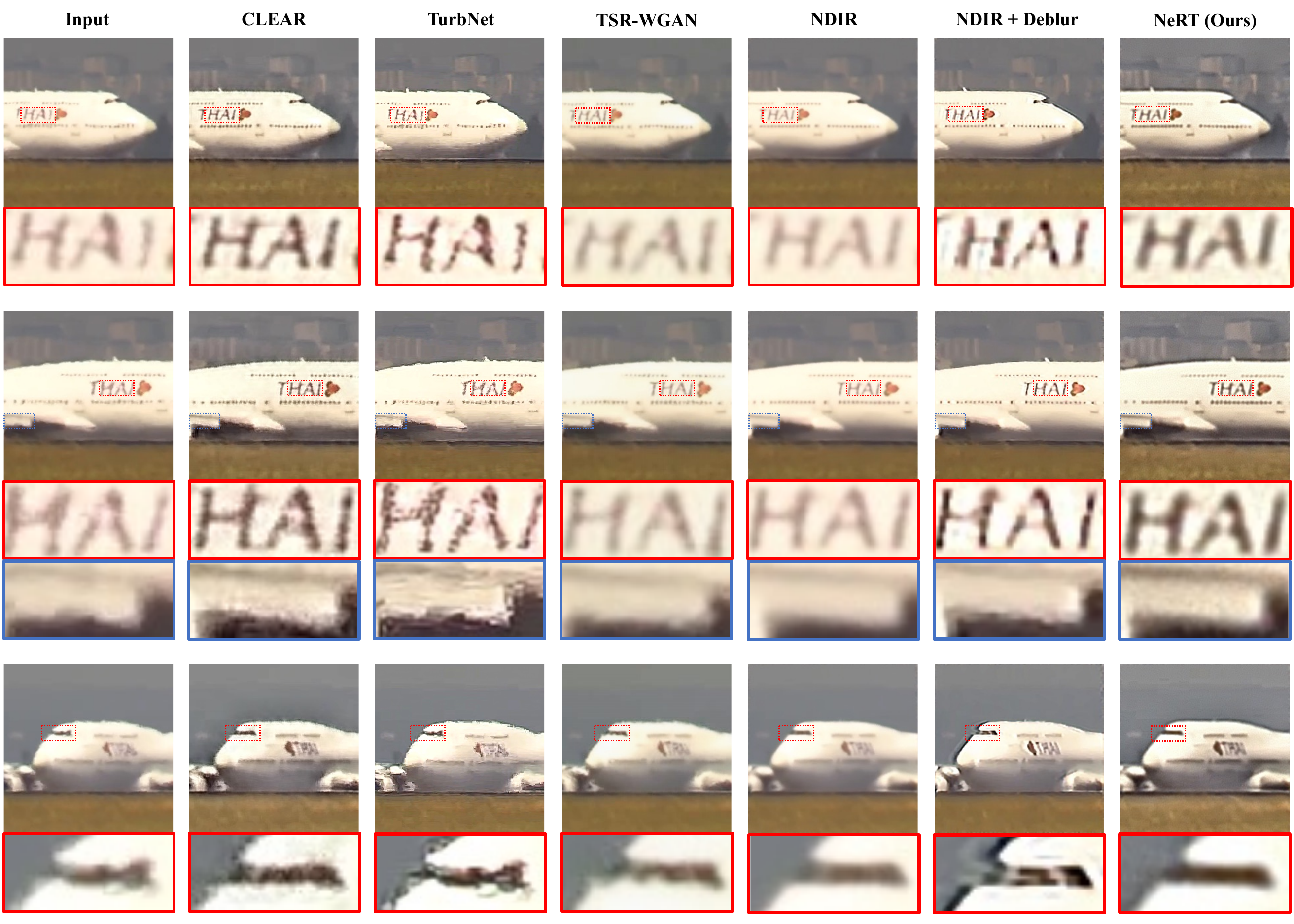} 
\caption{Qualitative results from the dynamic scene atmospheric turbulence dataset, airport dataset~\cite{anantrasirichai2018atmospheric}. CLEAR~\cite{anantrasirichai2018atmospheric} is a single-image reconstruction algorithm while other methods are multi-image reconstruction algorithms. All the input distorted images are fixed the same for all the multi-image reconstruction algorithms. Among those input distorted images, the sharpest one is chosen as the input for the single-image reconstruction algorithm TurbNet~\cite{zhang2022imaging}. We compare NeRT with other supervised~\cite{mao2022single, jin2021neutralizing} and unsupervised~\cite{li2021unsupervised, anantrasirichai2018atmospheric} SOTA. We are able to recover the undistorted and clean logo of "THAI" Airways while other methods show distorted and noisy logos. }
	\label{fig:dynamic}
\end{figure*}
\noindent \textbf{Shift-varying blurring} $\mathcal{P_\alpha}$ take generated tilted images $\mathbf{I}_\mathcal{T}  \in \mathbb{R}^{m \times n \times 3}$ as inputs and output generated tilted-then-blurred images $\mathbf{I}_{\mathcal{B} \circ \mathcal{T}}  \in \mathbb{R}^{m \times n \times 3}$. Shift-varying blurring leverages the Phase-to-Space (P2S) transform~\cite{mao2021accelerating} to apply pixel-wise spatially and temporally varying blurring. We initialize per-pixel correlated Zernike coefficients $\bm{\alpha} = [\alpha_1,\dots,\alpha_K]$ by multiplying independent and identically distributed Gaussian vectors with pre-computed correlation matrices. In addition, we apply P2S transform network during the optimization to convert Zernike coefficients $\bm{\alpha}= [\alpha_1,\dots,\alpha_K]$ to PSF basis coefficients $\bm{\beta} =[\beta_1,\dots,\beta_K]$. Together with the pre-computed PSF basis, we are able to use the converted basis coefficient $\bm{\beta}$ to compute spatially and temporally varying PSF for the shift-varying blurring operation. Note that we don't consider the first two Zernike bases since they are already accounted for as tilt by the grid deformers.

\subsection{Parameter initialization}
\begin{figure*}[!t]
	\centering
		\includegraphics[width=7.0in]{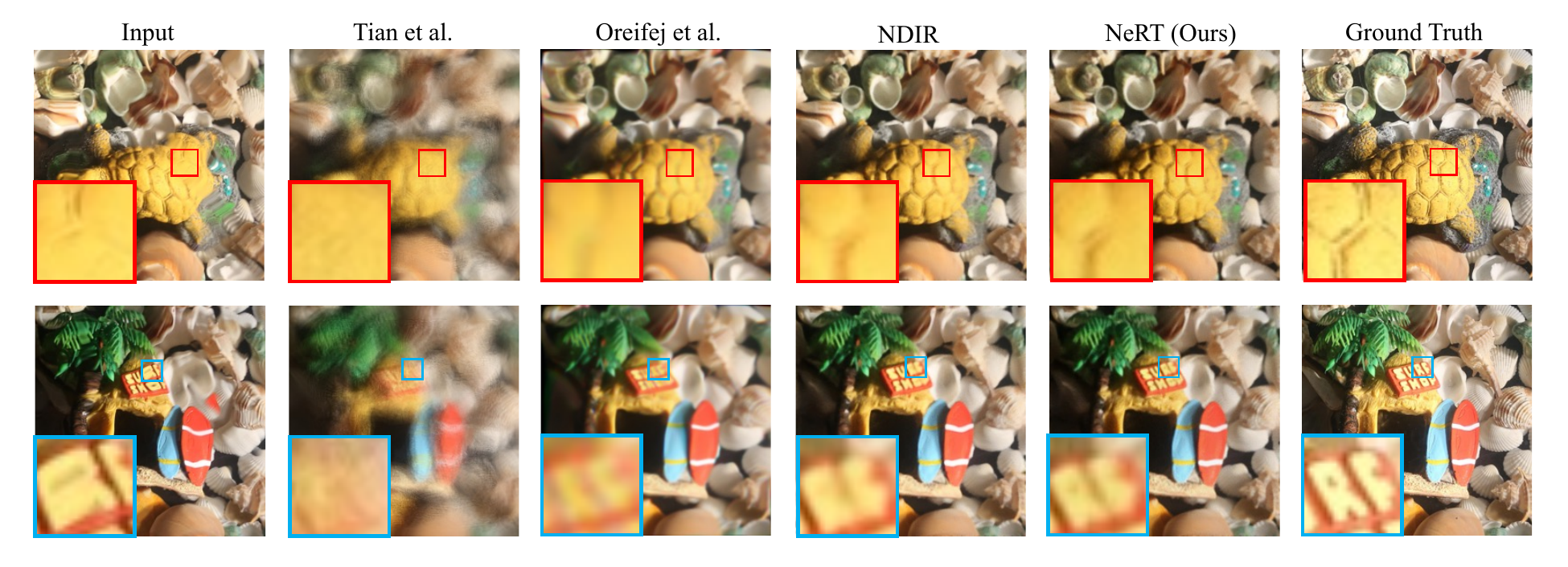} 
\caption{Qualitative results from the real water turbulence tank datasets~\cite{li2018learning}. We compare NeRT with other optimization-based water turbulence mitigation methods~\cite{tian2009seeing, oreifej2011two} and the state-of-the-art deep learning-based approaches~\cite{li2021unsupervised}. The qualitative results of NeRT are on par with SOTA performances. We highlight the best results using bold text.}
\label{fig:real_water}
\end{figure*}

\begin{figure*}[!t]
	\centering
		\includegraphics[width=7.0in]{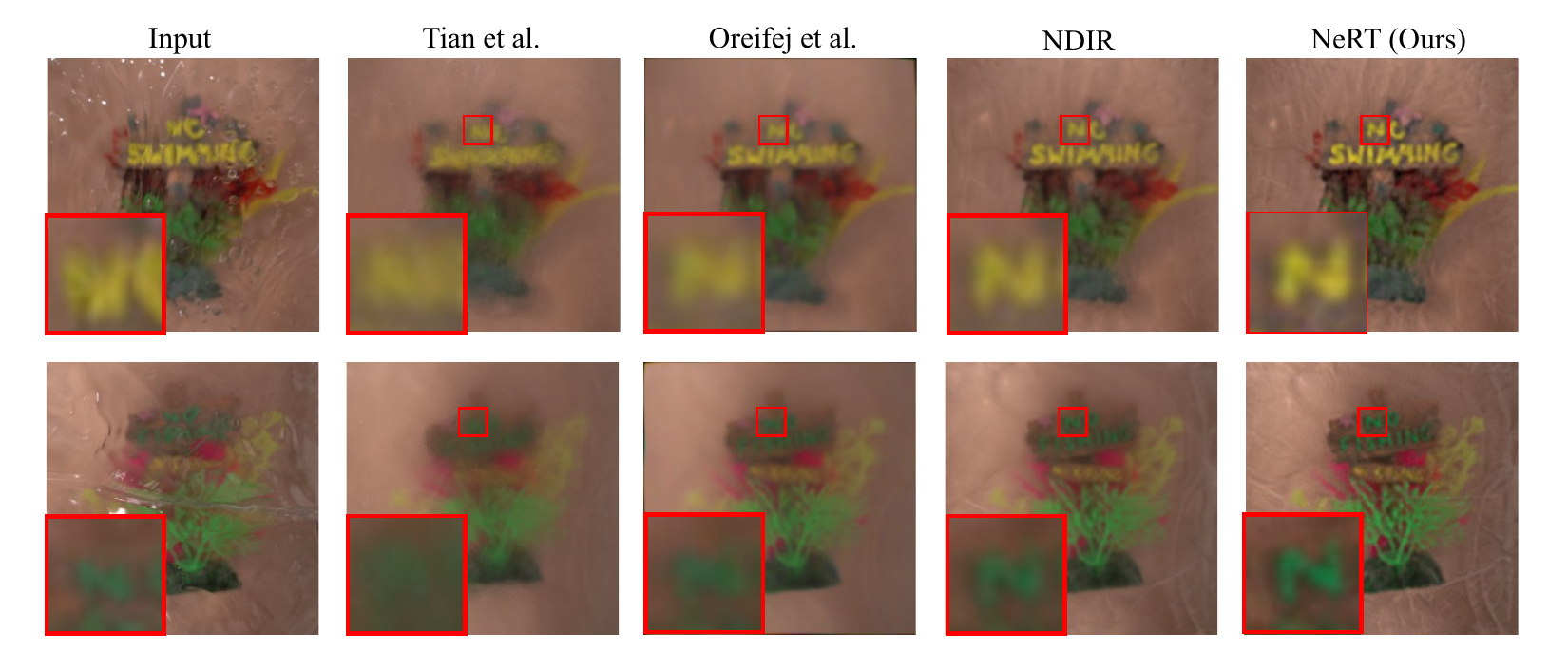} 
\caption{Qualitative results from our own custom mineral oil datasets. We compare NeRT with other optimization-based water turbulence mitigation methods~\cite{tian2009seeing, oreifej2011two} and the state-of-the-art deep learning-based approaches~\cite{li2021unsupervised}. The qualitative results of NeRT outperform SOTA performances. We highlight the best results using bold text.}
\label{fig:real_water}
\end{figure*}

\begin{figure*}[!h]
	\centering
		\includegraphics[width=7.0in]{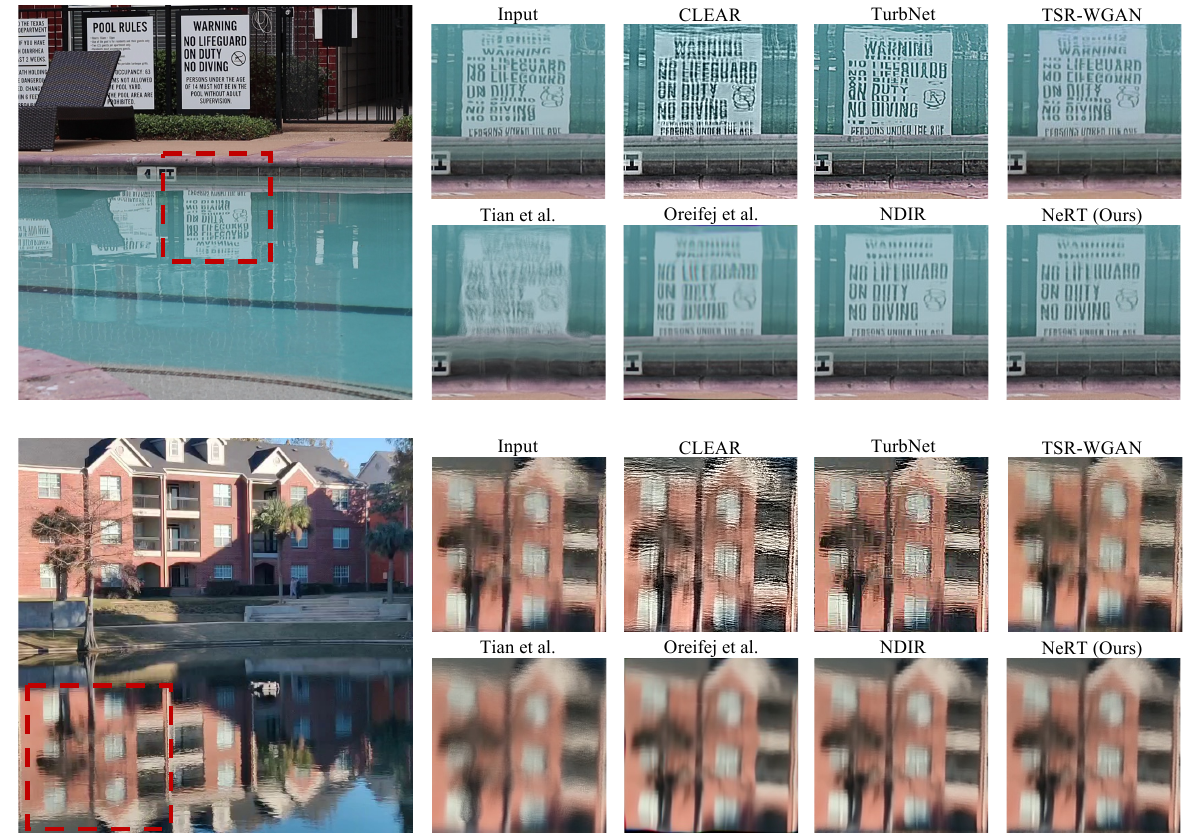} 
\caption{Qualitative results obtained from uncontrolled environments featuring water ripple reflection turbulence. We compare NeRT with other water and atmospheric turbulence mitigation methods. NeRT demonstrates the ability to effectively eliminate water ripple turbulence from reflections and produce clean images. The recovered images are shown mirrored for ease of viewing. }
\label{fig:wild_water}
\end{figure*}
Since our method is unsupervised, parameter initialization is rather important. A good initial point could help our model avoid saddle points and local minimums during the optimization. First, $D/r_0$, which characterizes the strength of the atmospheric turbulence and typically ranges from $1.0$ to $5.0$, determines the variance of the i.i.d. Gaussian vector during $\alpha$ initialization. Higher $D/r_0$ means stronger atmospheric turbulence. When the observed images have relatively high turbulence strength, our model converges better with a higher $D/r_0$. Second, $corr$~\cite{mao2021accelerating} refers to how correlated these nearby PSFs are and typically ranges from $-5$ to $-0.01$. A higher value means a stronger correlation. When the observed images have relatively high turbulence strength, our model converges better with a higher $corr$. Third, the kernel size of the PSF basis should vary as the image size varies. Large image dimensions usually require a large kernel size of the PSF basis.
\subsection{Two-step optimization}

We follow the network initialization in NDIR~\cite{li2021unsupervised}. During the first initialization step, the grid deformers $\mathcal{G}_{\theta}$ are constrained to learn an identity mapping from uniform grid $\mathbf{G_U}$ to be close to the uniform grid $\mathbf{G_U}$. In this way, we can limit the grid deformation from extreme pixel mixing. The image generator must learn an average of all the distorted input images. The loss function of the first initialization step is formulated as

\begin{equation}
    \min_{\theta,\phi} \sum_k \| \mathcal{G}_{\theta}^{k}(\mathbf{G}_U)\ - \mathbf{G}_U\|_1 + \| \mathcal{I}_{\phi}(\mathbf{G}_U) - \mathbf{I}_k \|_1.
\end{equation}
During the second iterative optimization step, we initialize the $\bm{\alpha}$ in shift-varying blurring, choosing appropriate $D/r_0$, $corr$, and PSF kernel size. The loss function is then formulated as
\begin{equation}
\begin{split}
    \min_{\theta, \phi, \alpha} \sum_k \|\mathcal{P}_{\alpha}( \mathcal{I}_{\phi}(\mathcal{G}_{\theta}^k(\mathbf{G}_U))) - \bm{I}_k \|_1 & \\
     +\| \mathcal{P}_{\alpha}( \mathbf{I}_\mathcal{T}^G) - \bm{I}_k \|_1 & \\
    +\|  \mathcal{P}_{\alpha}( \mathcal{I}_{\phi}(\mathcal{G}_{\theta}^k(\mathbf{G}_U))) - \mathcal{P}_{\alpha}( \mathbf{I}_\mathcal{T}^G) \|_1,
\end{split}
\label{eqn:consist_loss}
\end{equation}
where $\mathbf{I}_\mathcal{T}^G$ is a resampled tilted image given deformed grids $\mathbf{G}_\mathcal{T}$. We would like to enforce consistency between predicted tilted-then-blurred images, observed images, and resampled tilted-then-blurred images.

\section{Experiments and results}
\label{sec:exp}
In this section, we compare NeRT with other state-of-the-art supervised, unsupervised, and optimization-based methods, such as CLEAR~\cite{anantrasirichai2018atmospheric}, TurbNet~\cite{mao2022single}, TSR-WGAN~\cite{jin2021neutralizing} and NDIR~\cite{li2021unsupervised} on both real and synthetic atmospheric turbulence mitigation datasets. CLEAR~\cite{anantrasirichai2018atmospheric} is an optimization-based multi-frame restoration method. TurbNet~\cite{mao2022single} is a deep learning-based supervised single-frame restoration method. TSR-WGAN~\cite{jin2021neutralizing} is a deep learning-based supervised multi-frame restoration method. NDIR~\cite{li2021unsupervised} is a deep learning-based unsupervised multi-frame restoration method. We show that NeRT exhibits superior performance in both qualitative and quantitative assessments when compared to the state-of-the-art unsupervised and supervised approaches. 

Furthermore, we compare NeRT with other state-of-the-art optimization-based and deep learning-based water turbulence removal approaches, such as Tian et al.~\cite{tian2009seeing}, Oreifej et al.~\cite{oreifej2011two} and NDIR~\cite{li2021unsupervised}. Tian et al.~\cite{tian2009seeing} and Oreifej et al.~\cite{oreifej2011two} are optimization-based restoration methods. NDIR is the state-of-the-art deep learning-based unsupervised restoration method. We demonstrate the robustness of NeRT in addressing water turbulence mitigation problems. Specifically, both the qualitative and quantitative outcomes of NeRT are on par with the state-of-the-art methods.
\subsection{Implementation details}
We implement our model in Pytorch with one NVIDIA A100 80GB GPU. We use Adam~\cite{kingma2014adam} optimizer with a learning rate of $1\times 10^{-4}$ to update parameters in grid deformers $\mathcal{G}_{\theta}$, image generator $\mathcal{I}_{\phi}$ and shift-varying blurring $\mathcal{P}_{\alpha}$. We use 1000 epochs for the first initialization and the second iterative optimization steps, respectively. We empirically choose $D/r_0 = 5.0$, $corr = -5.0$, and a PSF kernel size of $11$ for all atmospheric turbulence mitigation experiments. For all the water turbulence mitigation experiments, we choose $D/r_0 = 0.1$, $corr = -5.0$, and a PSF kernel size of $5$. We resize all the distorted input images to $256 \times 256$. We randomly choose $20$ distorted images as input for all the experiments.
\subsection{Atmospheric turbulence mitigation}
\subsubsection{Evaluation on synthetic atmospheric turbulence datasets}

We chose images from mirflickr25k dataset~\cite{huiskes2008mir} as the ground truth images and used a simulator~\cite{mao2021accelerating} with three turbulence strengths ($D/r0 = 1.5$, $D/r0 = 3$ and $D/r0 = 4.5$). And we simulated 20 distorted images for each individual set of images and level of turbulence intensities. Table~\ref{tab:syn} demonstrates the quantitative comparison between our method, NeRT, and other SOTA. NeRT outperforms other methods in terms of peak signal-to-noise ratio (PSNR) and structural similarity index measure (SSIM)~\cite{wang2004image} and is robust to different turbulence strengths. 
\begin{table}[!t]
\centering
\caption{Quantitative results from the real water turbulence tank datasets~\cite{li2018learning}. $\uparrow$ means the higher the better. Results obtained using our method, NeRT, are on par with the current state-of-the-art NDIR~\cite{li2021unsupervised}.  We highlight the best and second-best results using bold and underlined
text, respectively.}
\begin{tabular}{|c|c|c|c|c|}
\hline
Method & \begin{tabular}[c]{@{}c@{}}Tian et al. \\ \cite{tian2009seeing}\end{tabular}  & \begin{tabular}[c]{@{}c@{}}Oreifej et al. \\ \cite{oreifej2011two}\end{tabular} & \begin{tabular}[c]{@{}c@{}}NDIR \\ \cite{li2021unsupervised}\end{tabular}            & NeRT (Ours)        \\ \thickhline
PSNR $\uparrow$    & $15.653$      & $17.031$        & $\mathbf{17.457}$ & $\underline{17.213}$ \\ \hline
SSIM $\uparrow$   & $0.349$      & $0.524 $  & $\mathbf{0.649}$ & $\underline{0.645}$       \\ \hline
\end{tabular}
\label{tab:real_water}
\end{table}
\subsubsection{Evaluation on real atmospheric turbulence datasets}

\noindent \textbf{Static scenes.} We include three real datasets for static scenes, the Siemens star dataset~\cite{gilles2017open}, the text dataset~\cite{mao2022single}, and the door dataset~\cite{gilles2017open}. Figure~\ref{fig:static} shows the qualitative results of these static scenes. NeRT achieves the best overall performance in terms of spatial resolution, high-contrast text reconstruction, and fine details recovery. Further, NeRT is able to preserve the fine details, such as the wire fences, while suppressing the blurring and tilting caused by atmospheric turbulence.

\noindent \textbf{Dynamic scenes.} We present the moving car dataset~\cite{anantrasirichai2018atmospheric} for dynamic scenes. The distorted input image sequences depict a car moving from back to front and from left to right. Figure~\ref{fig:dynamic} presents the qualitative results from the dynamic scene. Again we compare with other SOTA. NeRT is able to recover a high-contrast license plate with higher fidelity. To handle the dynamic scene, the image generator $\mathcal{I}_{\alpha}$ in our unsupervised model converges to a reference frame as a starting point during the first initialization step. During the second iterative optimization step, the clean image generated by the image generator $\mathcal{I}_{\alpha}$ is further optimized, given dozens of distorted images. 

\subsection{Seeing through water}
\label{sec:water}
We evaluate NeRT as well as other state-of-the-art methods on the real water turbulence dataset~\cite{li2018learning}. Li et al.~\cite{li2018learning} utilize a camera, agitating pump, and tank full of water to capture a real water turbulence tank dataset in their laboratory. This tank test dataset contains six different kinds of underwater scenes, including surf shops, dolphins, planes, ships, skulls, and turtles. We demonstrate the quantitative and qualitative results of NeRT in Table~\ref{tab:real_water} and Figure~\ref{fig:real_water}, respectively. We observe that the results obtained by NeRT are on par with the state-of-the-art performance.  

\section{Ablation studies}
We perform a series of ablation studies to validate our architecture design. Throughout the ablation studies, we use the physics-based atmospheric turbulence simulator~\cite{mao2021accelerating} to generate synthetic atmospheric turbulence datasets under different turbulence strengths ($D/r_0$).

\noindent \textbf{Number of distorted input images.} Table~\ref{tab:num_input} demonstrates the quantitative performance of NeRT with the different number of distorted input images. On one hand, we can observe that as the number of distorted input images increases, NeRT's performance improves gradually. On the other hand, we also observe the manifestation of the law of diminishing marginal utility. Specifically, as the quantity of distorted input images rises, NeRT's performance improvement rate decreases.
\begin{table}[]
\caption{Average PSNR, SSIM comparison with the different number of distorted input images. The number in the parentheses indicates the performance improvement rates. $\uparrow$ means the higher, the better.}
\begin{tabular}{|c|c|c|c|c|c|}
\hline
\begin{tabular}[c]{@{}c@{}}\# of \\ inputs\end{tabular}  & $2$      & $5$      & $10$     & $15$     & $20$     \\ \thickhline
PSNR                                                                    & $19.410$ & \begin{tabular}[c]{@{}c@{}}$20.076$ \\ $(+3.43\%)$\end{tabular} & \begin{tabular}[c]{@{}c@{}}$20.492$ \\ $(+2.07\%)$\end{tabular}  & \begin{tabular}[c]{@{}c@{}}$20.627$\\ $(+0.66\%)$\end{tabular} & \begin{tabular}[c]{@{}c@{}}$20.712$\\ $(+0.41\%)$\end{tabular} \\ \hline
SSIM                                                                    & $0.601$  & \begin{tabular}[c]{@{}c@{}}$0.639$\\ $(+6.32\%)$\end{tabular}  & \begin{tabular}[c]{@{}c@{}}$0.655$\\ $(+2.50\%)$\end{tabular}  & \begin{tabular}[c]{@{}c@{}}$0.660$\\ $(+0.76\%)$\end{tabular}  & \begin{tabular}[c]{@{}c@{}}$0.664$\\ $(+0.61\%)$\end{tabular}  \\ \hline
\end{tabular}
\label{tab:num_input}
\end{table}
\begin{figure*}[!ht]
	\centering
		\includegraphics[width=\linewidth]{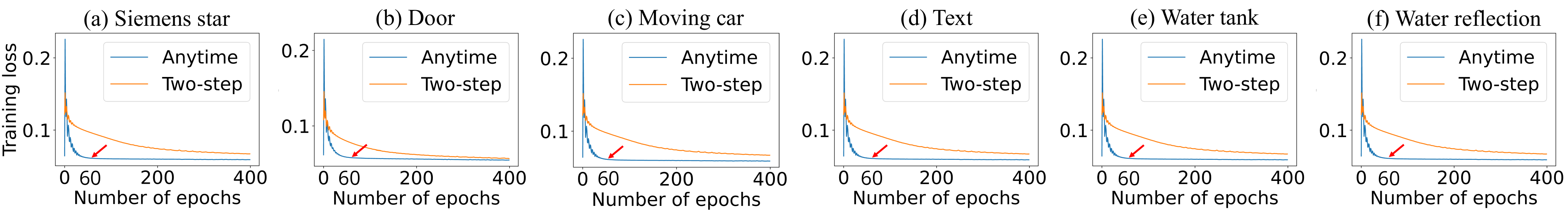} 
\caption{NeRT converges $\sim 48\times$ faster after two-step optimization (initialization) over multiple datasets: (a) Siemens star dataset~\cite{gilles2017open}, (b) door dataset~\cite{gilles2017open}, (c) moving car dataset~\cite{anantrasirichai2018atmospheric}, (d) text dataset~\cite{mao2022single}, and (e) water tank dataset~\cite{li2018learning}, and (f) water ripple reflection dataset. It takes a total of $2000$ epochs ($\sim 8$ minutes) to converge during the two-step optimization stage, while it only takes $60$ epochs ($\sim 10$ seconds) to converge after the initialization (red mark).}
	\label{fig:anytime}
\end{figure*}

\noindent \textbf{Consistency between $\mathbf{I}_{\mathcal{B}\circ \mathcal{T}}$, $\mathbf{I}^{G}_{\mathcal{B}\circ \mathcal{T}}$, and $\mathbf{I}$.}  we also conducted thorough ablation studies to analyze the significance of consistency between predicted tilted-then-blurred images $\mathbf{I}_{\mathcal{B}\circ \mathcal{T}}$, resampled tilted-then-blurred images $\mathbf{I}^{G}_{\mathcal{B}\circ \mathcal{T}}$, and observed images $\mathbf{I}$ in~\eqref{eqn:consist_loss}). Table~\ref{tab:consis} exhibits the effect of different consistency losses on the performance of NeRT. 
\begin{table}[!ht]
\caption{Average PSNR, SSIM comparison with different consistency losses in~\eqref{eqn:consist_loss}. $Loss_1$ stands for the $L_1$ loss between predicted tilted-then-blurred images $\mathbf{I}_{\mathcal{B}\circ \mathcal{T}}$ and observed image $\mathbf{I}$. $Loss_2$ stands for the $L_1$ loss between resampled tilted-then-blurred images $\mathbf{I}^{G}_{\mathcal{B}\circ \mathcal{T}}$ and observed image $\mathbf{I}$. $Loss_3$ stands for the $L_1$ loss between resampled tilted-then-blurred images $\mathbf{I}^{G}_{\mathcal{B}\circ \mathcal{T}}$ and predicted tilted-then-blurred images $\mathbf{I}_{\mathcal{B}\circ \mathcal{T}}$.}
\begin{tabular}{|c|c|c|c|c|}
\hline
Method & \begin{tabular}[c]{@{}c@{}}NeRT \\ w/o $Loss_1$\end{tabular} & \begin{tabular}[c]{@{}c@{}}NeRT \\ w/o $Loss_2$\end{tabular} & \begin{tabular}[c]{@{}c@{}}NeRT \\ w/o $Loss_3$\end{tabular} & \begin{tabular}[c]{@{}c@{}}NeRT \\ (Proposed)\end{tabular} \\ \hline
PSNR   & 20.075                                                 & 20.035                                                & 20.350                                                & 20.712                                                     \\ \hline
SSIM   & 0.629                                          & 0.628                                                       & 0.649                                                       & 0.664                                              \\ \hline
\end{tabular}
\label{tab:consis}
\end{table}

\section{Water ripple turbulence mitigation in the wild}
\label{sec:wild}
To showcase the versatility of NeRT in addressing a wide range of turbulence mitigation challenges, we venture into uncontrolled environments and capture a couple of real-life video recordings featuring water ripple turbulence. These recordings comprise both clean and undistorted scenes, as well as their corresponding mirror images that are distorted by water ripple turbulence. We apply our method to the parts of the video that are distorted by the water ripples and reconstruct clear images.
Figure~\ref{fig:wild_water} exhibits some qualitative results of mitigating the water ripple turbulence in the wild.

\section{Toward live reconstructions of continuous video frames}
\label{sec:anytime}
One drawback of unsupervised implicit neural representation methods is that they are iterative and have a slower reconstruction time. The parameters of the network need to be optimized for every new scene. Here, we show a path of extending our method toward live reconstruction by processing the current video frame based on network parameters learned from previous frames. This speeds up the convergence time, leading to what we call the ``anytime" reconstruction.

Our method NeRT, like NDIR~\cite{li2021unsupervised}, has a separate grid deformer $\mathcal{G_{\theta}}$ and a separate shift-varying blurring $\mathcal{P_{\alpha}}$ for each distorted input image, and shares a single image generator $\mathcal{I_{\phi}}$ across all the distorted images. During anytime reconstruction, we initialize a new separate grid deformer $\mathcal{G_{\theta}}$ and shift-varying blurring $\mathcal{P_{\alpha}}$ for the most recent frame that is captured. All the other grid deformers $\mathcal{G_{\theta}}$, shift-varying blurring $\mathcal{P_{\alpha}}$ and the image generator $\mathcal{I_{\phi}}$ can contain all the information of the previously observed distorted frame, speeding up the anytime reconstruction.

Figure~\ref{fig:anytime} demonstrates $48\times$ speedup of our method during anytime convergence. It takes about $8$ minutes to complete the two-step initialization step for the first set of frames. However, converging for every newly captured video frame only takes about $10$ seconds.
\section{Conclusions and discussions}
We have proposed the first unsupervised and physically grounded model for general turbulence mitigation. Given multiple observed distorted images, our model leveraged the physically correct tilt-then-blur model to reconstruct a clean and undistorted image. Our model could generalize and outperform other SOTA in various scenarios, such as seeing through water, seeing through turbulence, and mitigating water reflection turbulence. After the two-step initialization, our method converged $48 \times$ faster on the latest captured frame.

\noindent \textbf{Limitations and future directions.} Our shift-varying deblurring did not have any regularization. Thus, the reconstructed clean image inevitably consisted of some noise due to blind deconvolution. A more sophisticated shift-varying deblurring process remains a future research direction. Additionally, one might also leverage the implicit neural network for image superresolution. As we know, the implicit neural network is a continuous representation of the image. More pixel coordinates queried into the implicit image function lead to higher resolution images generated.

\bibliographystyle{IEEEtran}
\bibliography{TCI_manuscript}
\vfill

\end{document}